\documentclass{article}
\usepackage{spconf,amsmath,graphicx,hyperref}
\usepackage{multirow}
\usepackage{cite}
\usepackage{amssymb,amsfonts}
\usepackage{algorithmic}
\usepackage{textcomp}
\usepackage{xcolor}
\usepackage{enumitem}

\title{MiLorE-SSL: Scaling Multilingual Capabilities in Self-Supervised Models without Forgetting}
\name{Jing Xu, Minglin Wu, Xueyuan Chen, Xixin Wu, Helen Meng}
\address{The Chinese University of Hong Kong}
\begin{document}
\ninept
\maketitle
\begin{abstract}
Self-supervised learning (SSL) has greatly advanced speech representation learning, but multilingual SSL models remain constrained to languages encountered during pretraining. Retraining from scratch to incorporate new languages is computationally expensive, while sequential training without migitation strategies often leads to catastrophic forgetting. To address this, we propose MiLorE-SSL, a lightweight framework that combines LoRA modules with a soft mixture-of-experts (MoE) mechanism for efficient continual multilingual training. LoRA provides efficient low-rank adaptation, while soft MoE promotes flexible expert sharing across languages, reducing cross-lingual interference. 
To further mitigate forgetting, we introduce limited replay data from existing languages, avoiding reliance on large historical corpora. 
Experiments on ML-SUPERB demonstrate that MiLorE-SSL achieves strong performance in new languages and improves the ability in existing ones with only 2.14\% trainable parameters.
\end{abstract}
\section{Introduction}
Self-supervised learning (SSL) has become the cornerstone of speech representation learning, with models such as HuBERT \cite{hubert}, Wav2vec 2.0 \cite{wav2vec}, and WavLM \cite{wavlm} achieving strong performance across tasks such as automatic speech recognition (ASR), speech translation, and speech synthesis. By learning universal representations from large-scale unlabeled audio, these models effectively capture both acoustic and semantic features.

Despite these advances, extending SSL models to support multiple languages remains challenging. Multilingual SSL models such as XLSR\cite{xlsr}, mHuBERT\cite{mhubert}, mHuBERT-147\cite{mhubert147}, and MMS\cite{scalingmms} have broadened coverage to dozens of languages. However, retraining such models from scratch whenever new languages emerge is computationally expensive and impractical given the scale of global linguistic diversity. 
To address the language imbalance between high-resource and low-resource languages, works such as mHuBERT-147 and MMS have adopted data up-sampling strategies to artificially increase the presence of underrepresented languages. 
While this enhances representation quality for low-resource languages, it introduces significant data redundancy and training inefficiency. 
High-resource language data must be down-sampled, while low-resource data are repeatedly sampled to maintain balance, resulting in uneven utilization and increased training cycles. 
More critically, this static balancing scheme requires complete dataset reprocessing and full retraining when new languages are added, making it rigid and costly. These limitations highlight the need for continual training, enabling models to incrementally learn new languages without discarding prior knowledge or incurring prohibitive computational costs.

However, sequential training without mitigation strategies often leads to catastrophic forgetting \cite{overcomingcf}, where performance on previously learned languages severely degrades when new languages are introduced. To mitigate this, prior approaches have explored replay-based strategies \cite{replay} and adapter-style modules \cite{parameter_continual}. Replay-based strategies\cite{replay} alleviate forgetting by reusing historical data but often require large memory storage and access to historical corpora. Adapter-style modules\cite{parameter_continual} introduce additional modules for new tasks or languages, but their rigid design scales poorly when the number of languages grows. 
More recently, parameter-efficient finetuning techniques such as LoRA\cite{lora} and DoRA\cite{dora} have shown promise by updating only small low-rank modules while keeping the backbone frozen. For example, \cite{seamlesslanguage} demonstrated that LoRA can extend HuBERT to Mandarin with reduced forgetting. 
However, these methods still face limitations in representational capacity and scalability, as cross-lingual interference increases as the number of languages grows.

Meanwhile, mixture-of-experts (MoE) architectures \cite{firstmoe} have emerged as a powerful paradigm for scaling model capacity without proportionally increasing computational cost. Unlike dense models that activate all parameters for every input, MoE selectively activates expert modules according to the routing strategy.  In Natural Language Processing (NLP), works such as GShard\cite{gshard} and SwitchTransformers\cite{switchformer}, and recent large-scale LLMs \cite{moe,glam} demonstrate that sparsely activated experts can specialize across tasks and languages while maintaining efficient inference. GShard introduced sparsely gated MoE layers into Transformer architectures, achieving significant performance gains in multilingual translation. 
SwithTransformers further simplifies routing by activating only one expert per token, reducing memory and communication overhead while preserving model quality. In the speech domain, models\cite{speechinformed,mole,speechconformermoe} have shown improvements in multilingual ASR by routing inputs to different experts. Despite these advances, applying MoE to extend language coverage of SSL models remains underexplored. 
Most existing MoE models are trained in a static setting, where all languages are known in advance, limiting adaptability when new languages emerge. Moreover, hard routing strategies may lead to unstable expert assignments, whereas soft routing allows more flexible sharing.

In this work, we propose MiLorE-SSL (\textbf{Mi}xture of \textbf{Lor}A \textbf{E}xperts), the first framework that combines LoRA with soft MoE for efficient continual training of SSL models. LoRA provides efficient low-rank adaptation, while soft MoE flexibly routes inputs to multiple experts, promoting knowledge sharing between languages and reducing interference. To further mitigate forgetting, we integrate a replay strategy using limited samples from existing languages, avoiding reliance on full historical corpora. Our contributions are:
\begin{figure*}
\centering
\vspace{-0.6em}
\begin{minipage}[b]{0.3\linewidth}
  \centering
  \centerline{\includegraphics[width=2in]{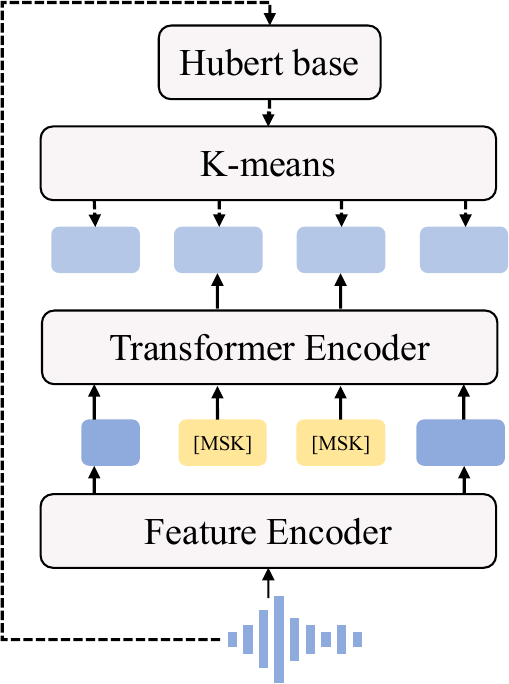}}
  \centerline{(a)}\medskip
\end{minipage}
\begin{minipage}[b]{0.3\linewidth}
  \centering
  \centerline{\includegraphics[width=2in]{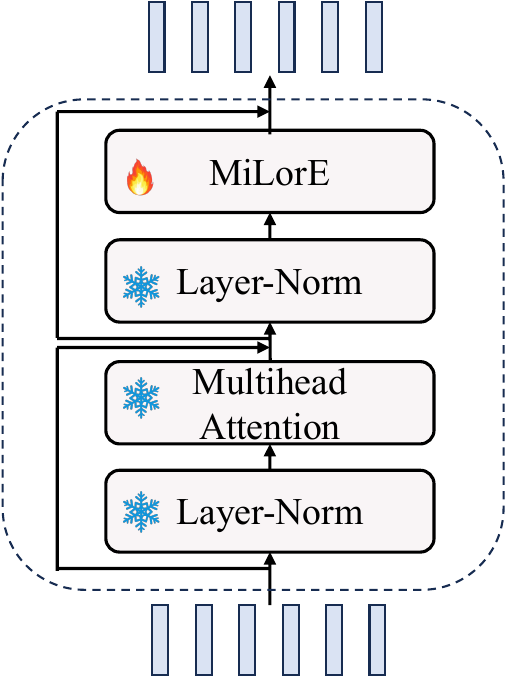}}
  \centerline{(b)}\medskip
\end{minipage}
\begin{minipage}[b]{0.3\linewidth}
  \centering
  \centerline{\includegraphics[width=2in]{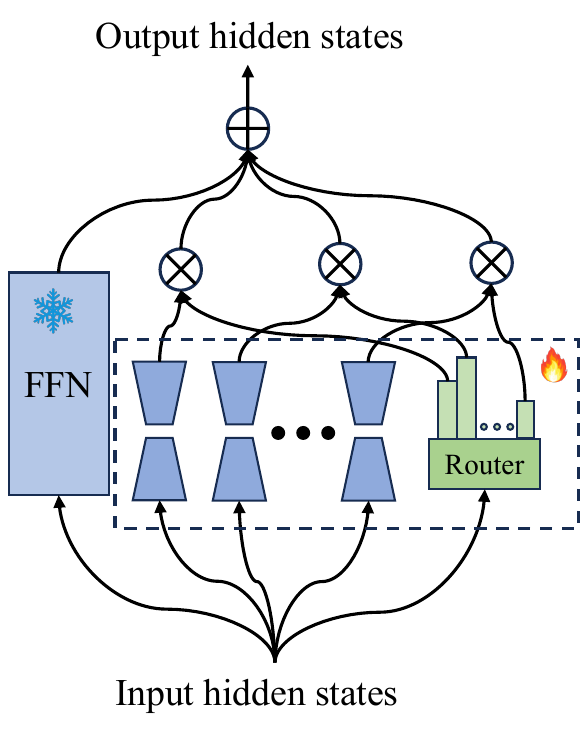}}
  \centerline{(c)}\medskip
\end{minipage}
%
\vspace{-1em}
\caption{Overview of MiLorE-SSL framework. (a) Architecture of HuBERT-based SSL models. (b) Transformer block with MiLorE module. (c) Architecture of MiLorE module, where a router selects experts to process input hidden states alongside a frozen FFN.}

\label{fig:milore_ssl_arch}
\vspace{-1em}
\end{figure*}
\begin{itemize}
\item We propose MiLorE-SSL, the first framework combining LoRA and soft MoE to extend the language coverage of SSL models.
\item We design a soft expert routing mechanism that balances knowledge sharing and specialization across languages.
\item We incorporate replay with minimal existing-language data to reduce forgetting without access to full historical data.
\item Experiments on ML-SUPERB benchmarks demonstrate that MiLorE-SSL achieves high performance on new languages and improves performance on existing ones using only 2.14\% trainable parameters.
\end{itemize}

\section{Methodology}
\label{sec:method}

This section presents the design of our proposed MiLorE-SSL framework for continual multilingual extension of self-supervised speech models. We first describe the architecture and core components of MiLorE-SSL in Section~\ref{sec:milore-ssl}, followed by the replay strategy used to mitigate catastrophic forgetting in Section~\ref{sec:replay}. Finally, we detail the overall training procedure in Section~\ref{sec:training strategy}.

\subsection{MiLorE-SSL architecture}
\label{sec:milore-ssl}
As illustrated in Figure~\ref{fig:milore_ssl_arch} (a), HuBERT-based SSL models consist of a feature encoder and a Transformer network. The Transformer network is composed of $L$ blocks, each containing a multi-head self-attention layer and a feed-forward network (FFN). To enable efficient multilingual extension, we modify each Transformer block by replacing its feed-forward network with a MiLorE module, as shown in Figure~\ref{fig:milore_ssl_arch} (b)(c). The resulting model augmented with MiLorE modules is referred to as MiLorE-SSL.

Each MiLorE module comprises three components: (1) the frozen FFN backbone with weights $W_0$; (2) a set of LoRA-based experts ${E_1,E_2,...,E_N}$; and (3) a soft router that assigns input-dependent weights to experts. Importantly, the FFN backbone and other components of Transformer blocks remain frozen to preserve prior multilingual knowledge. The LoRA experts introduce lightweight trainable low-rank matrices, extending the model capacity to new languages at low cost. The router dynamically computes soft routing weights, enabling flexible knowledge sharing and specialization across languages without requiring language identifiers. 

Given an input hidden state $h_{in}$, the output of MiLorE module is computed as:
\begin{align}
    o = W_0h_{in} + \sum_{i=1}^{N} p_i\cdot E_i(h_{in})
\end{align}
where $W_0 \in \mathbb{R}^{d_{out} \times d_{in}}$ is the frozen FFN weight matrix, and $p_i$ are the soft routing weights computed via:
\begin{align}
p=softmax(W_rh_{in})
\end{align}
where $W_r$ is the trainable weight matrix of the router. Specifically, each expert $E_i$ is parameterized using LoRA:
\begin{align}
\Delta W_{E_i} = B_iA_i
\end{align}
where matrix $B_i \in \mathbb{R}^{d_{out} \times r}$, $A_i \in \mathbb{R}^{r\times d_{in}}$, and low-rank dimension $r\ll min(d_{in},d_{out})$.
Thus, the overall output of the MiLorE module becomes:
\begin{align}
    o 
    = W_0h_{in} + \sum_{i=1}^{N} p_i \cdot B_iA_ih_{in}
\end{align}
Compared to traditional MoE designs, where each expert is a dense feed-forward network, MiLorE substantially reduces parameter cost while still retaining expert specialization. By integrating MiLorE modules into SSL models, MiLorE-SSL combines LoRA's efficiency with MoE's specialization, enabling effective and scalable multilingual extension with minimal forgetting.
\subsection{Replay strategy}
\label{sec:replay}
To further mitigate catastrophic forgetting during continual training, we adopt a replay strategy. A small subset of samples from previously learned languages is included during continual training. This strategy allows the model to revisit prior language distributions, reinforcing previously acquired knowledge while adapting to new linguistic patterns. This strategy also mitigates the storage and computational burden associated with full-data replay, making it practical for scalable multilingual learning.

\subsection{Training strategy}
\label{sec:training strategy}
The lightweight and modular design of MiLorE-SSL makes it suitable for scalable language extension by updating only the LoRA-based experts and the router in MilorE-SSL modules. Moreover, target labels for continual training can be obtained from existing SSL models, avoiding the need for the costly two-stage training. 

The continual training objective remains consistent with the original HuBERT's masked prediction loss:
\begin{align}
L = -\sum_{t\in \mathcal{M}} log P(z_t|\hat{h}_t)
\end{align}
where $z_t$ is the target cluster index obtained via K-means over SSL representations, and $\hat{h}_t$ is the contextual representation from the Transformer encoder, and $\mathcal{M}$ is the set of masked frames.

During continual training, the loss is jointly applied to new-language samples and replayed samples. This ensures the model not only learns new linguistic features, but also preserves knowledge of previously learned ones, resulting in stable and effective continual multilingual extension.

\section{Experimental Setup}
\label{sec:setup}
\subsection{Datasets}
We follow the data selection pipeline of mHuBERT-147\cite{mhubert147}, using its provided manifests\footnote{\url{https://huggingface.co/utter-project/mHuBERT-147-base-3rd-iter/tree/main/manifest}} to extract Mandarin (zh-CN) and Cantonese (zh-HK, yue) data from CommonVoice 11.0\cite{commonvoice}. For Mandarin, we also include Thchs-30\cite{thchs30}, AISHELL-1\cite{aishell}, and AISHELL-3\cite{aishell3}, consistent with mHuBERT-147. However, we exclude VoxLingua107, leading to a reduced Mandarin training set. To implement the replay strategy, we randomly sample 100 hours of English data from CommonVoice 11.0. Given the lightweight nature of MiLorE-SSL, no data up-sampling strategy is applied.

After filtering utterances outside the range of [2,30] seconds, the final training data distribution is summarized in Table\ref{tab: training data statistics}. Specifically, the final dataset includes 105.072 hours of Cantonese (same as mHuBERT-147), 341.867 hours of Mandarin (smaller than 382.7 hours in mHuBERT-147), and 100.555 hours of English (much smaller than 47,210.7 hours in mHuBERT-147). This setup enables us to evaluate the ability of MiLorE-SSL to extend to new languages while mitigating forgetting.

For evaluation, we consider monolingual ASR and language identification (LID). Monolingual ASR is evaluated on CommonVoice (in-domain) and Fleurs\cite{fleurs} (out-of-domain) corpora. For LID, we mix one-hour subsets from both datasets and remove long utterances.
\begin{table}[ht]
\centering
\vspace{-1em}
\caption{Statistics of training data (filtered)}
\label{tab: training data statistics}
\resizebox{\linewidth}{!}{
\begin{tabular}{lccccc}
\hline
\multicolumn{1}{c}{Language}  & data source     &   \# sentences & total samples &  \# hours & total hours\\ 
\hline
\multirow{4}{*}{Mandarin(cmn)} 
 & Thchs-30 & 10000 &\multirow{4}{*}{265767}& 25.545  &\multirow{4}{*}{341.867} \\
 &AISHELL-1 & 120188& &151.147\\
 &AISHELL-3 & 58223&&60.607 \\
 &CommonVoice\_zh-CN &77356&&104.568\\ \hline
\multirow{2}{*}{Cantonese (yue)}
 &    CommonVoice\_zh-HK  & 78641 &\multirow{2}{*}{92435}& 89.503&\multirow{2}{*}{105.072}\\
& CommonVoice\_yue & 13794& &15.569\\\hline
\multirow{1}{*}{English (eng)}
&CommonVoice\_en  &  74203 & \multirow{1}{*}{74203}& 105.555& \multirow{1}{*}{100.555}\\
\hline
\end{tabular}
}
\vspace{-0.6em}
\end{table}
\subsection{Training configurations}
We continually train HuBERT-Large to enhance its capability on Mandarin and Cantonese. Training targets are derived from K-means clustering on features from the 9th layer of mHuBERT-147. The model is trained for 400k steps with a peak learning rate of $1.5\times10^{-3}$. MiLorE-SSL is configured with two LoRA experts, each with a rank of 12, resulting in only 2.14\% trainable parameters.

Clustering is performed via MiniBatchKMeans with a batch size of 10k frames and K-means++ initialization over 20 seeds. For each language, 50 hours of speech are randomly sampled for clustering.

\subsection{Evaluation configuration}
We evaluate our model on the ML-SUPERB benchmark, focusing on monolingual ASR and LID tasks under the 1-hour fine-tuning setting. ASR performance is measured by Character Error Rate (CER), while LID is evaluated by Accuracy (ACC).  All evaluations follow ESPnet recipes\footnote{\url{https://github.com/espnet/espnet/tree/master/egs2/ml_superb/asr1}}, with a learning rate of $1\times10^{-4}$. To stabilize training under limited data, we select intermediate representations as input: for monolingual ASR, the 22nd layer of HuBERT-Large and the 11th layer of HuBERT-Base are used; for LID, the 21st layer of HuBERT-Large and the 10th layer of HuBERT-Base are used.

We compare MiLorE-SSL against:
\begin{itemize}
\item \textbf{mHuBERT-147}: A multilingual HuBERT model pretrained on 90,000 hours of speech from 147 languages, serving as both reference and target label extractor.

\item \textbf{HuBERT-large}: An English HuBERT model pretrained on 60,000 hours of Libri-Light, adopted as the base model for continual training. 
\end{itemize}
\section{Experimental results}
\label{sec:results}
\begin{table*}
\centering
\caption{Performance comparison across ASR and LID tasks on CommonVoice and Fleurs. Bold marks the best, "avg" denotes the average.}
\label{tab: main results}
\resizebox{0.95\linewidth}{!}{
\begin{tabular}{lcccccccccccc}
\hline
\multirow{3}{*}{System}   & \multicolumn{8}{c}{Monolingual ASR (CER$\downarrow$)} & \multicolumn{4}{c}{\multirow{2}{*}{LID (ACC$\uparrow$)}} \\
\cline{2-9}
& \multicolumn{4}{c}{CommonVoice}    & \multicolumn{4}{c}{Fleurs}  \\ 
\cline{2-13}
&eng & cmn & yue & avg & eng &cmn &yue & avg & eng &cmn & yue & avg\\
\hline\hline
mHuBERT-147 (1iter) &30.2 & 24.7 & 21.8 & 25.57 & 26.6 & 25.6 & 25.3 & 25.83 & 93.41 & 91.30 &93.10 &92.60\\
mHuBERT-147 (2iter)   & 21.2 & 17.4 & 15.8 & 18.13 & 18.2 & 16.4 & 17.2 & 17.27 & 97.60 & 96.89 & 98.85& 97.78\\
mHuBERT-147 (3iter)    &18.5 & 15.5 & 14.8 & 16.27 & 15.9 &15.1 & 15.6 &15.53 & 98.20 & 96.89 & 96.55 & 97.21\\
HuBERT\_Large  & 11.5 & 21.2 & 17.6& 16.77 & 10.4 & 21.2 & 18.1  & 16.57 & 97.60 & 90.68 & 96.55 & 94.94\\
 \hline

\textbf{MiLorE-SSL (Ours)}   & \textbf{10.3} & 10.7& 11.0 & \textbf{10.67} & \textbf{9.4} & 10.2 & 11.6 & \textbf{10.40} & \textbf{99.40} & \textbf{99.38} & \textbf{99.43} & \textbf{99.40}\\
 \hspace{1em}-MoE  & 11.0 & 11.3 & 11.1 &11.13 & 10.0 & 10.7 &11.9 &10.87 & 98.20 & 98.76 & \textbf{99.43} & 98.80\\
 \hspace{1em}-Replay & 26.6 & 10.8& 10.9 &16.10 & 24.8 & 10.3 & \textbf{11.4} & 15.50 & 98.20 & 97.52 & \textbf{99.43} & 98.28\\
 \hspace{1em}-MoE+Replay &27.5 & \textbf{10.6} & \textbf{10.6} & 16.23 &24.4 &\textbf{9.9} &11.5 & 15.27& 95.21 & 88.82 & \textbf{99.43} & 94.49\\

 \hline
\end{tabular}
}
\end{table*}

\subsection{Main results}
Table~\ref{tab: main results} presents a comprehensive comparison of ASR and LID performance across English, Mandarin, and Cantonese, evaluated on both CommonVoice (in-domain) and Fleurs (out-of-domain).

From Table~\ref{tab: main results}, we observe that HuBERT\_Large, pretrained solely on English, performs strongly on English ASR, achieving 11.5\% CER on CommonVoice and 10.4\% on Fleurs. These results outperform the best mHuBERT-147 model, which obtains 18.5\% and 15.9\% CER, respectively. However, HuBERT\_Large struggles on non-English languages, with CERs of 21.2\% (Mandarin) and 17.6\% (Cantonese) on CommonVoice, substantially worse than 15.5\% and 14.8\% for mHuBERT-147. While mHuBERT-147 achieves more balanced multilingual performance and benefits from additional training iterations, its overall performance still lags behind HuBERT\_Large, especially on English. This gap highlights the need for continual multilingual extension of HuBERT\_Large. 

In contrast, MiLorE-SSL consistently improves performance across all languages and tasks. On CommonVoice, MiLorE-SSL reduces CERs to 10.3\% (English), 10.7\% (Mandarin), and 11.0\% (Cantonese). This demonstrates substantial gains over mHuBERT-147 (18.5\%, 15.5\%, 14.8\%) and HuBERT\_Large (11.5\%, 21.2\%, 17.6\%). Similar improvements are observed on Fleurs, confirming that MiLorE-SSL generalizes well to out-of-domain data. For LID, MiLorE-SSL attains an average accuracy of 99.40\%, outperforming both mHuBERT-147 (97.21\%) and HuBERT\_Large (94.94\%).  

These results demonstrate that MiLorE-SSL enables parameter-efficient continual training, effectively extending HuBERT\_Large to new languages while retaining its strong English ability. Remarkably, this is achieved with only 100 hours of replay data, showing that our method effectively mitigates forgetting without requiring extensive historical data. Besides, since HuBERT-large is trained on LibriLight, our replay strategy does not require access to historical data and avoids storing or sub-sampling the original training corpus.

\subsection{Analysis of Expert activation patterns}
\begin{figure}[ht]
\centering
\vspace{-0.6em}
\includegraphics[width=0.98\linewidth]{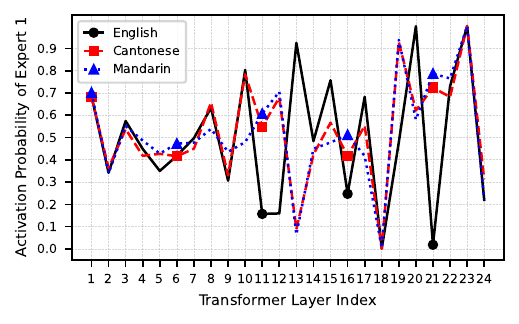}
%
\vspace{-1.6em}
\caption{Layer-wise expert weights across languages }
\label{fig:f1_weights}
 \vspace{-1em}
\end{figure}
Figure~\ref{fig:f1_weights} visualizes the activation probabilities of the first expert across Transformer layers for English, Cantonese, and Mandarin. As each layer contains only two experts, the probability of the first expert effectively reflects the overall pattern. 

In the lower layers, all three languages exhibit nearly identical activation patterns, indicating that these layers encode language-independent features, consistent with prior findings that lower layers primarily capture speaker-related information. In the middle layers (6-10), Cantonese aligns more with English, while at layers 5, 13, 14, 19, and 20, it resembles Mandarin. Layers 15-17 show strong divergence across all three languages, reflecting language-specific representations. 

These observations highlight the value of introducing MoE structure in multilingual modeling, as it enables flexible modeling of both shared and language-specific features. This is further validated by the ablation results in Table~\ref{tab: main results}, where removing the MoE component consistently degrades performance across all languages and tasks, confirming its critical role in MiLorE-SSL.  

\subsection{Ablation Study}

We analyze the impact of LoRA rank and the number of experts. As shown in Tables~\ref{tab: same trainable} and \ref{tab: varied experts}, increasing LoRA rank under a fixed number of experts generally improves performance, as each expert gains stronger modeling capacity. Conversely, increasing the number of experts under a fixed rank also helps, likely due to better language separation. However, with the same limited trainable parameters, too many experts degrade performance, likely due to reduced expert capacity and unstable routing.  

Table~\ref{tab: main results} also shows that removing MoE architecture consistently harms all tasks, confirming its role in balancing shared and language-specific representations. Without MoE, the model reduces to a pure LoRA setup (rank 24) like ~\cite{seamlesslanguage}, with identical trainable parameters to MiLorE-SSL. The performance drop highlights the importance of expert specialization. Besides, removing replay strategy primarily affects English performance, underscoring its role in mitigating forgetting. Combining MoE and Replay yields the best results, offering both representational flexibility and training stability. 

\begin{table}
\centering
\vspace{-0.6em}
\caption{Effect of expert count under fixed trainable parameters}
\label{tab: same trainable}
\resizebox{0.96\linewidth}{!}{
\begin{tabular}{cccccccccc}
\hline
\multirow{2}{*}{lora rank}  & \multirow{2}{*}{\# experts} & \multicolumn{4}{c}{CommonVoice}    &\multicolumn{4}{c}{Fleurs}  \\ 
\cline{3-10}& &eng & cmn & yue & avg & eng &cmn &yue & avg\\
\hline
 12 & 2 & \textbf{10.3} & \textbf{10.7}& 11.0 & \textbf{10.67} & 9.4 & \textbf{10.2} & \textbf{11.6} & \textbf{10.40}\\
8 & 3  & 10.6 &11.6 & \textbf{10.8} &11.00 & 9.7 & 10.3  & 12.3  & 10.77 \\
6 & 4  & 10.6  & 11.2 & 11.2 & 11.00 & 9.5 & 10.5 & 11.7 & 10.57 \\
4 & 6 & 10.4 &11.5 & 11.8  &  11.23& \textbf{9.3} & 10.8 & 12.7 &  10.93\\
3 & 8  & 10.8 & 11.8 &11.6  & 11.40 &  9.5 &  11.3 & 12.2 & 11.00 \\
2 & 12  & 10.5 & 12.3 & 11.8 & 11.53& 9.5 & 11.3 & 12.5 & 11.10\\
\hline
\end{tabular}
}
\vspace{-1em}
\end{table}
\begin{table}
\centering
\caption{Impact of LoRA rank and expert count on ASR task}
\label{tab: varied experts}
\resizebox{0.96\linewidth}{!}{
\begin{tabular}{cccccccccc}
\hline
\multirow{2}{*}{lora rank}  & \multirow{2}{*}{\# experts} & \multicolumn{4}{c}{CommonVoice}    &\multicolumn{4}{c}{Fleurs}  \\ 
\cline{3-10}& &eng & cmn & yue & avg & eng &cmn &yue & avg\\
\hline
12 & 2 & 10.3 & 10.7& 11.0 & 10.67 & 9.4 & 10.2 & 11.6 & 10.40\\
12 & 3  & \textbf{10.2} & 11.2& \textbf{10.5} & 10.63 & 9.1 &  \textbf{9.8}& 11.5 & 10.13\\ 
12 & 4  & 10.4  & \textbf{10.6} & 10.6 & \textbf{10.53} &  \textbf{9.0}& \textbf{9.8} & \textbf{11.1} & \textbf{9.97}\\
\hline
16 & 2  & 10.3 & 10.7& 10.7 & 10.57 & 9.5 & 10.0 & 11.7 &10.40 \\
8 & 2  & 10.3 &11.5 & 11.4 & 11.07& 9.6 & 10.8 & 11.9 & 10.77\\
4&2& 10.4& 11.9& 12.1 &11.47 &9.9 & 12.0 &12.6 &11.50\\
\hline
\end{tabular}
}
\vspace{-1em}
\end{table}
\section{Conclusion}

    We introduced MiLorE-SSL, a parameter efficient framework for continual multilingual learning of self-supervised speech models. 
    By combining low-rank adaptation with a soft mixture-of-experts architecture, MiLorE-SSL effectively extends HuBERT\_Large to new languages while preserving performance on existing ones. To further mitigate forgetting, we incorporate a replay strategy that revisits a small subset of data from previously learned language without access to large historical datasets. Experimental results show consistent improvements over HuBERT\_Large, particularly in Mandarin and Cantonese, without compromising English capability. In the future, we plan to explore on more languages and exploit the layer-wise information hierarchy of SSL models to design more effective architectures.

\section{Acknowledgement}
This study was supported by Human-Computer Communications Laboratory  (HCCL), Department of Systems Engineering and Engineering Management, The Chinese University of Hong Kong, Hong Kong SAR, China and the Centre for Perceptual and Interactive Intelligence (CPII) Ltd., a CUHK-led InnoCentre under the InnoHK initiative of the Innovation and Technology Commission of the Hong Kong Special Administrative Region Government.

\begingroup
\small
\bibliographystyle{IEEEbib}
\bibliography{strings,refs}

@article{wav2vec,
  title={wav2vec 2.0: A framework for self-supervised learning of speech representations},
  author={Baevski, Alexei and Zhou, Yuhao and others},
  journal={Advances in neural information processing systems},
  volume={33},
  pages={12449--12460},
  year={2020}
}

@article{hubert,
  title={Hubert: Self-supervised speech representation learning by masked prediction of hidden units},
  author={Hsu, Wei-Ning and Bolte, Benjamin and others},
  journal={IEEE/ACM Transactions on Audio, Speech, and Language Processing},
  volume={29},
  pages={3451--3460},
  year={2021},
  publisher={IEEE}
}

@article{wavlm,
  title={Wavlm: Large-scale self-supervised pre-training for full stack speech processing},
  author={Chen, Sanyuan and Wang, Chengyi and Chen, Zhengyang and others},
  journal={IEEE Journal of Selected Topics in Signal Processing},
  volume={16},
  number={6},
  pages={1505--1518},
  year={2022},
  publisher={IEEE}
}

@article{xlsr,
  title={Unsupervised cross-lingual representation learning for speech recognition},
  author={Conneau, Alexis and Baevski, Alexei and others},
  journal={arXiv preprint arXiv:2006.13979},
  year={2020}
}

@article{overcomingcf,
  title={Overcoming catastrophic forgetting in neural networks},
  author={Kirkpatrick, James and Pascanu, Razvan and Rabinowitz and others},
  journal={Proceedings of the national academy of sciences},
  volume={114},
  number={13},
  pages={3521--3526},
  year={2017},
  publisher={National Acad Sciences}
}

@article{mhubert,
  title={Textless speech-to-speech translation on real data},
  author={Lee, Ann and Gong, Hongyu and Duquenne, Paul-Ambroise and others},
  journal={arXiv preprint arXiv:2112.08352},
  year={2021}
}

@article{lora,
  title={Lora: Low-rank adaptation of large language models},
  author={Hu, Edward J and Shen, Yelong and Wallis, Phillip and Allen-Zhu, Zeyuan and Li, Yuanzhi and Wang, Shean and Wang, Lu and Chen, Weizhu},
  journal={arXiv preprint arXiv:2106.09685},
  year={2021}
}

@article{replay,
  title={Experience replay for continual learning},
  author={Rolnick, David and Ahuja, Arun and Schwarz, Jonathan and Lillicrap, Timothy and Wayne, Gregory},
  journal={Advances in neural information processing systems},
  volume={32},
  year={2019}
}

@article{moe,
  title={Outrageously large neural networks: The sparsely-gated mixture-of-experts layer},
  author={Shazeer, Noam and Mirhoseini, Azalia and Maziarz, Krzysztof and Davis, Andy and Le, Quoc and Hinton, Geoffrey and Dean, Jeff},
  journal={arXiv preprint arXiv:1701.06538},
  year={2017}
}

@inproceedings{seamlesslanguage,
  title     = {Seamless Language Expansion: Enhancing Multilingual Mastery in Self-Supervised Models},
  author    = {Jing Xu and Minglin Wu and Xixin Wu and Helen Meng},
  year      = {2024},
  booktitle = {Interspeech 2024},
  pages     = {4973--4977},
  doi       = {10.21437/Interspeech.2024-1716},
  issn      = {2958-1796},
}

@article{thchs30,
  title={Thchs-30: A free chinese speech corpus},
  author={Wang, Dong and Zhang, Xuewei},
  journal={arXiv preprint arXiv:1512.01882},
  year={2015}
}

@article{firstmoe,
  title={Adaptive Mixtures of Local Experts},
  author={Robert A. Jacobs and Michael I. Jordan and Steven J. Nowlan and Geoffrey E. Hinton},
  journal={Neural Computation},
  year={1991},
  volume={3},
  pages={79-87},
  url={https://api.semanticscholar.org/CorpusID:572361}
}

@inproceedings{glam,
  title={Glam: Efficient scaling of language models with mixture-of-experts},
  author={Du, Nan and Huang, Yanping and Dai, Andrew M and others},
  booktitle={International conference on machine learning},
  pages={5547--5569},
  year={2022},
  organization={PMLR}
}

@inproceedings{speechinformed,
  title={Mixture of informed experts for multilingual speech recognition},
  author={Gaur, Neeraj and Farris, Brian and Haghani, Parisa and Leal, Isabel and Moreno, Pedro J and Prasad, Manasa and Ramabhadran, Bhuvana and Zhu, Yun},
  booktitle={ICASSP 2021-2021 IEEE International Conference on Acoustics, Speech and Signal Processing (ICASSP)},
  pages={6234--6238},
  year={2021},
  organization={IEEE}
}

@inproceedings{mole,
  title={Mole: Mixture of language experts for multi-lingual automatic speech recognition},
  author={Kwon, Yoohwan and Chung, Soo-Whan},
  booktitle={ICASSP 2023-2023 IEEE International Conference on Acoustics, Speech and Signal Processing (ICASSP)},
  pages={1--5},
  year={2023},
  organization={IEEE}
}

@article{speechconformermoe,
  title={Mixture-of-expert conformer for streaming multilingual asr},
  author={Hu, Ke and Li, Bo and Sainath, Tara N and Zhang, Yu and Beaufays, Francoise},
  journal={arXiv preprint arXiv:2305.15663},
  year={2023}
}

@inproceedings{aishell3,
  title     = {AISHELL-3: A Multi-Speaker Mandarin TTS Corpus},
  author    = {Yao Shi and Hui Bu and Xin Xu and Shaoji Zhang and Ming Li},
  year      = {2021},
  booktitle = {Interspeech 2021},
  pages     = {2756--2760},
  doi       = {10.21437/Interspeech.2021-755},
  issn      = {2958-1796},
}

@inproceedings{aishell,
  title={Aishell-1: An open-source mandarin speech corpus and a speech recognition baseline},
  author={Bu, Hui and Du, Jiayu and Na, Xingyu and Wu, Bengu and Zheng, Hao},
  booktitle={O-COCOSDA 2017},
  pages={1--5},
  year={2017},
  organization={IEEE}
}

@article{commonvoice,
  title={Common voice: A massively-multilingual speech corpus},
  author={Ardila, Rosana and Branson, Megan and Davis, Kelly and others},
  journal={arXiv preprint arXiv:1912.06670},
  year={2019}
}

@article{switchformer,
  title={Switch transformers: Scaling to trillion parameter models with simple and efficient sparsity},
  author={Fedus, William and Zoph, Barret and Shazeer, Noam},
  journal={Journal of Machine Learning Research},
  volume={23},
  number={120},
  pages={1--39},
  year={2022}
}

@inproceedings{dora,
  title={Dora: Weight-decomposed low-rank adaptation},
  author={Liu, Shih-Yang and Wang, Chien-Yi and Yin, Hongxu and Molchanov, Pavlo and Wang, Yu-Chiang Frank and Cheng, Kwang-Ting and Chen, Min-Hung},
  booktitle={Forty-first International Conference on Machine Learning},
  year={2024}
}

@article{scalingmms,
  title={Scaling speech technology to 1,000+ languages},
  author={Pratap, Vineel and Tjandra, Andros and others},
  journal={Journal of Machine Learning Research},
  volume={25},
  number={97},
  pages={1--52},
  year={2024}
}

@article{gshard,
  title={Gshard: Scaling giant models with conditional computation and automatic sharding},
  author={Lepikhin, Dmitry and Lee, HyoukJoong and Xu, Yuanzhong and Chen, Dehao and Firat, Orhan and Huang, Yanping and Krikun, Maxim and Shazeer, Noam and Chen, Zhifeng},
  journal={arXiv preprint arXiv:2006.16668},
  year={2020}
}

@inproceedings{parameter_continual,
  title={Parameter-efficient transfer learning for NLP},
  author={Houlsby, Neil and Giurgiu, Andrei and Jastrzebski, Stanislaw and Morrone, Bruna and De Laroussilhe, Quentin and Gesmundo, Andrea and Attariyan, Mona and Gelly, Sylvain},
  booktitle={International conference on machine learning},
  pages={2790--2799},
  year={2019},
  organization={PMLR}
}

@inproceedings{mhubert147,
  title     = {mHuBERT-147: A Compact Multilingual HuBERT Model},
  author    = {Marcely {Zanon Boito} and Vivek Iyer and others},
  year      = {2024},
  booktitle = {Interspeech 2024},
  pages     = {3939--3943},
  doi       = {10.21437/Interspeech.2024-938},
  issn      = {2958-1796},
}

@inproceedings{fleurs,
  title={Fleurs: Few-shot learning evaluation of universal representations of speech},
  author={Conneau, Alexis and Ma, Min and Khanuja, Simran and Zhang, Yu and Axelrod, Vera and Dalmia, Siddharth and Riesa, Jason and Rivera, Clara and Bapna, Ankur},
  booktitle={2022 IEEE Spoken Language Technology Workshop (SLT)},
  pages={798--805},
  year={2023},
  organization={IEEE}
}
\endgroup

\end{document}